\documentclass[runningheads]{llncs}
\usepackage[T1]{fontenc}
\usepackage{graphicx}
\usepackage{multirow}
\usepackage{multicol}
\usepackage{tikz} 
\usepackage[]{framed}
\usepackage{sectsty}
\usepackage{amsmath}
\usepackage{graphicx}
\usepackage[colorlinks=true, allcolors=blue]{hyperref}
\usepackage{booktabs}
\usepackage{adjustbox}
\usepackage{xspace}
\usepackage{soulutf8}
\usepackage{placeins}  
\usepackage{tcolorbox}
\tcbuselibrary{breakable}
\usepackage{float}
\newcommand\stwoorc{\textsc{s2orc-doc2json}\xspace}
\newcommand\multixscience{\textsc{Multi-XScience}\xspace}
\newcommand\multinews{\textsc{Multi-News}\xspace}
\newcommand\multilexsum{\textsc{Multi-LexSum}\xspace}
\newcommand\howsumm{\textsc{HowSumm}\xspace}
\newcommand\findsum{\textsc{FINDSum}\xspace}
\newcommand\bigsurvey{\textsc{BigSurvey}\xspace}
\newcommand\sumpubmed{\textsc{SumPubMed}\xspace}
\newcommand\scisummnet{\textsc{ScisummNet}\xspace}
\newcommand\proposeddataset{\textsc{SurveySum}\xspace}
\newcommand\specter{\textsc{SPECTER2}\xspace}
\newcommand\checkeval{\textsc{Check-Eval}\xspace}

\tcbset{
    style_prompt/.style={
        fontlower=\tiny,
        fontupper=\small
    }
}

\begin{document}
\title{\proposeddataset: A Dataset for Summarizing Multiple Scientific Articles into a Survey Section}
\titlerunning{\proposeddataset: A Dataset for MDS}
%

\author{
Leandro Carísio Fernandes $^*$ \inst{1}\orcidID{0000-0002-4114-2334} \and
Gustavo Bartz Guedes $^*$ \inst{2}\orcidID{0000-0002-3278-9961} \and
Thiago Soares Laitz $^*$ \inst{2,3}\orcidID{0000-0001-7205-2094} \and
Thales Sales Almeida $^*$ \inst{2,3}\orcidID{0009-0006-9568-9331} \and
Rodrigo Nogueira \inst{2, 3}\orcidID{0000-0002-2600-6035} \and
Roberto Lotufo \inst{2, 4}\orcidID{0000-0002-5652-0852} \and
Jayr Pereira \inst{2, 5}\orcidID{0000-0001-5478-438X}
}

\authorrunning{L. C. Fernandes et al.}
%
\institute{
Instituto de Pesquisa para Economia Digital, Brasília-DF, Brasil, \email{carisio@gmail.com} \and
UNICAMP, Campinas-SP, Brasil \and
Maritaca AI, Campinas-SP, Brasil \and
NeuralMind.ai, Campinas-SP, Brasil \and
Universidade Federal do Cariri, Juazeiro do Norte-CE, Brasil, \email{jayr.pereira@ufca.edu.br}}
\maketitle              
\def\thefootnote{*}\footnotetext{Equal contribution}
\def\thefootnote{\arabic{footnote}}

\begin{abstract}
Document summarization is a task to shorten texts into concise and informative summaries. This paper introduces a novel dataset designed for summarizing multiple scientific articles into a section of a survey. Our contributions are: (1) \proposeddataset, a new dataset addressing the gap in domain-specific summarization tools; (2) two specific pipelines to summarize scientific articles into a section of a survey; and (3) the evaluation of these pipelines using multiple metrics to compare their performance. Our results highlight the importance of high-quality retrieval stages and the impact of different configurations on the quality of generated summaries.

\end{abstract}
\section{Introduction}

Document summarization is a field of artificial intelligence aimed at creating concise and informative summaries from extensive texts \cite{automatic_summarization}. This process allows users to quickly grasp the main information of a document without needing to read the entire original content. Summarization can be performed in two main ways: extractive and abstractive \cite{automatic_summarization}. Extractive summarization selects and copies key parts of the original text to compose the summary, preserving the verbatim of the source text. On the other hand, abstractive summarization rewrites the content, generating a summary that may include paraphrasing and synthesizing information more creatively and flexibly.

Multi-document summarization (MDS) extends these concepts to handle multiple texts simultaneously \cite{automatic_summarization}. The goal is to generate a single summary that incorporates information from a set of documents, eliminating redundancies and highlighting essential cross-referenced information. This technique is useful in situations where information on a specific topic is spread across various documents.

MDS datasets advance MDS research by providing essential data for training, testing, and validating models. These datasets support the development of algorithms capable of handling the complexities of summarizing multiple documents, enabling researchers to compare different summarization methods effectively \cite{Koh_2022}. Essentially, an MDS dataset is composed of summaries generated from multiple source documents. The composition of this dataset can be done in different ways, such as by selecting a set of documents and asking humans to write summaries, or by using existing summaries and their source documents. The latter is the case for the \multinews dataset \cite{fabbri2019multinews}, which uses news articles and their summaries to create an MDS dataset.

A comprehensive survey is a review that provides a broad overview of a specific topic, covering the most relevant and recent research in the field. These surveys are essential for researchers to understand the state-of-the-art and identify gaps in the literature. The automatic generation of a survey's text can be considered a MDS task, where a set of scientific articles is summarized to produce a cohesive text. This task is challenging due to the complexity of scientific literature and the need to maintain the technical accuracy of the original content. However, most available summarization datasets are designed with a general-purpose approach, covering a broad range of topics and document types. There is a lack of datasets dedicated to the task of summarizing multiple scientific articles to generate a survey's text, which poses a substantial barrier to advancements in this specialized area of research. This type of dataset assists the development and benchmarking of summarization models focusing on the generation of scientific surveys.

This paper introduces \proposeddataset\footnote{\url{https://huggingface.co/datasets/unicamp-dl/SurveySum}}, a dataset designed for the task of summarizing multiple scientific articles into survey sections. Our contribution includes: (1) the \proposeddataset dataset and the methodology to create it, which aims to address the lack of in domain-specific summarization tools; (2) two specific pipelines to summarize scientific articles into a section of a survey; and (3) evaluation of these pipelines in respect to consistency, quality, and relevance of the generated summaries.

\section{Related Work}


Multi-document summarization (MDS) addresses the problem of extracting information that is spread across multiple documents, making it more challenging than single-document summarization. It is still an evolving field, with no single approach to solve this problem. The development of this field relies on datasets that serve as benchmarks for evaluating and comparing different summarization methods. The scientific literature includes some examples of such datasets, each varying in domain, structure, size, and summarization objective. These datasets are important for advancing research and improving summarization models.


Outside the scientific domain, \multinews \cite{fabbri2019multinews} presents a large volume of news articles and summaries, focusing on the abstractive summarization of multiple documents in the journalistic context. Also in this context, Ghalandari et al. \cite{ghalandari2020largescale} proposed a large-scale dataset for multi-document summarization that contains concise human-written summaries of news events. In the legal field, \multilexsum tackles the summarization of legal cases, presenting a set of civil rights litigation summaries with multiple granularities. The summarization of instructional content is the focus of the \howsumm dataset, derived from WikiHow articles \cite{boni2021howsumm}. The \findsum dataset focuses on the challenge of summarizing long text and multiple tables \cite{liu2023long}.

In the scientific domain, the need to manage the vast amount of literature has generated interest in datasets specific to the summarization of scientific articles. \multixscience \cite{lu2020multixscience} is a dataset focused on the generation of ``related work'' sections by summarizing multiple scientific articles. \bigsurvey \cite{LIU_2022} was designed to create structured summaries of academic articles, focusing on consolidating literature reviews. \sumpubmed \cite{SumPubMed} and \scisummnet \cite{ScisummNet} aim the summarization of single scientific articles.

Several other datasets can be used for multi-document summarization. Koh et al. \cite{Koh_2022} work is an extensive survey on long document summarization, including ten other MDS datasets. 

However, to the best of our knowledge, there is no dataset designed to summarize a set of scientific articles in order to generate a section of a scientific survey. The closest dataset, \multixscience \cite{lu2020multixscience}, focuses on writing the ``related work'' section of a paper based on the abstract of the paper and the abstract of the articles it references. It does not consider the text of the referenced articles. Our dataset, on the other hand, adresses this gap: it considers the full text of the referenced articles to write a specific section on a subject.  Consequently, our proposed dataset is truly a multi-document summarization dataset, capable of handling lengthy documents.

\section{Method}

In this section, we describe the methodology to create \proposeddataset. First, we describe our approach in selecting existing surveys from the scientific literature. Later, we detail the text and citations extraction process.

\subsection{Survey Selection}

The selection of surveys was based on the authors' expertise in the field of artificial intelligence, natural language processing, and machine learning. The surveys were chosen to cover a broad range of topics within these areas, ensuring the dataset's diversity and applicability. 
For surveys to be eligible for inclusion in the dataset, they had to attend to the following criteria: (1) be a comprehensive survey in the field of artificial intelligence, natural language processing, or machine learning; (2) be divided into sections, each with citations of scientific articles; (3) be freely available online; and (4) be written in English.

First, we conducted a search on Google Scholar using the keywords \textit{comprehensive}, \textit{survey}, \textit{artificial intelligence}, \textit{natural language processing}, and \textit{machine learning}. We then filtered the results to include only surveys published in the last five years and that are freely available. The surveys were manually reviewed to ensure they met the eligibility criteria, resulting in the selection of six surveys.

Usually, surveys are structured into sections to organize the content and facilitate reading and understanding. This makes each section a self-contained summary of a specific aspect of the survey, while the cited papers are the source documents. Thus, each section is a summary of the cited papers, making it an ideal target for the MDS task. 

Finally, \proposeddataset samples are comprised by the sections texts of surveys with the corresponding citation papers. This ensures that the dataset can be used to evaluate the summarization of specific topic areas within the surveys.


\subsection{Dataset Creation}

\begin{figure}
    \centering
    \includegraphics[width=1\linewidth]{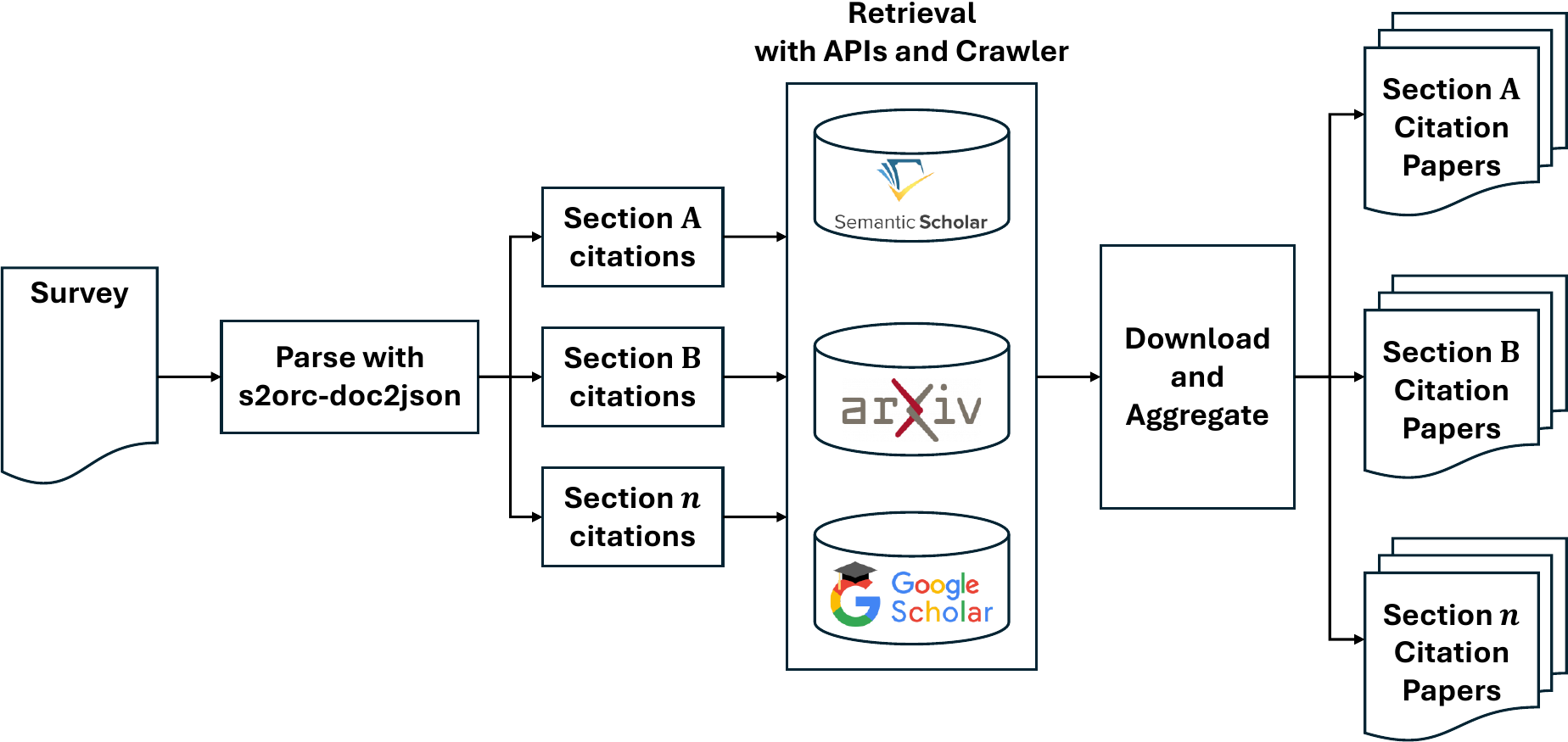}
    \caption{Dataset Creation Process.}
    \label{fig:dataset_creation}
\end{figure}

In this section, we describe the methodology to create the dataset, illustrated in Figure \ref{fig:dataset_creation}. This process is applied to all the surveys selected in the method described in the previous section.

Initially, we parsed each survey with \stwoorc\footnote{\url{https://github.com/allenai/s2orc-doc2json}}, extracting the sections along with the corresponding citations. \stwoorc converts scientific articles from various formats, such as PDFs and LaTeX, to JavaScript Object Notation (JSON) format. Additionally, \stwoorc extracts the citations from scientific papers, indicating both the position of the citation within the text and the referenced paper. 

The next step is to retrieve the full text of the cited papers. For this, we conducted searches, utilizing available APIs (e.g., arXiv search API), across online repositories including arXiv\footnote{\url{https://info.arxiv.org/help/api/index.html}} and Semantic Scholar\footnote{\url{https://www.semanticscholar.org/product/api}} supplemented by a crawler for Google Scholar\footnote{\url{https://scholar.google.com.br}}.

Finally, we manually verified each extracted citation against the original surveys. This step was necessary because \stwoorc is a machine learning based process, thus susceptible to inconsistencies. Also, some corresponding papers were not retrieved using the APIs or the crawler, so a manual search and retrieval was necessary.

Using this approach, the \proposeddataset was created. Data was extracted from 79 sections across 6 surveys. Each section is related to an average of 7.38 articles, with a standard deviation of 4.95 articles (median is 7 articles). Thus, the task is to generate a survey's section text given its title and a collection of papers.

\section{Pipelines}\label{sec:pipelines}

\subsection{General pipeline}

\begin{figure}[htbp]
    \centering
    \includegraphics[width=1\linewidth]{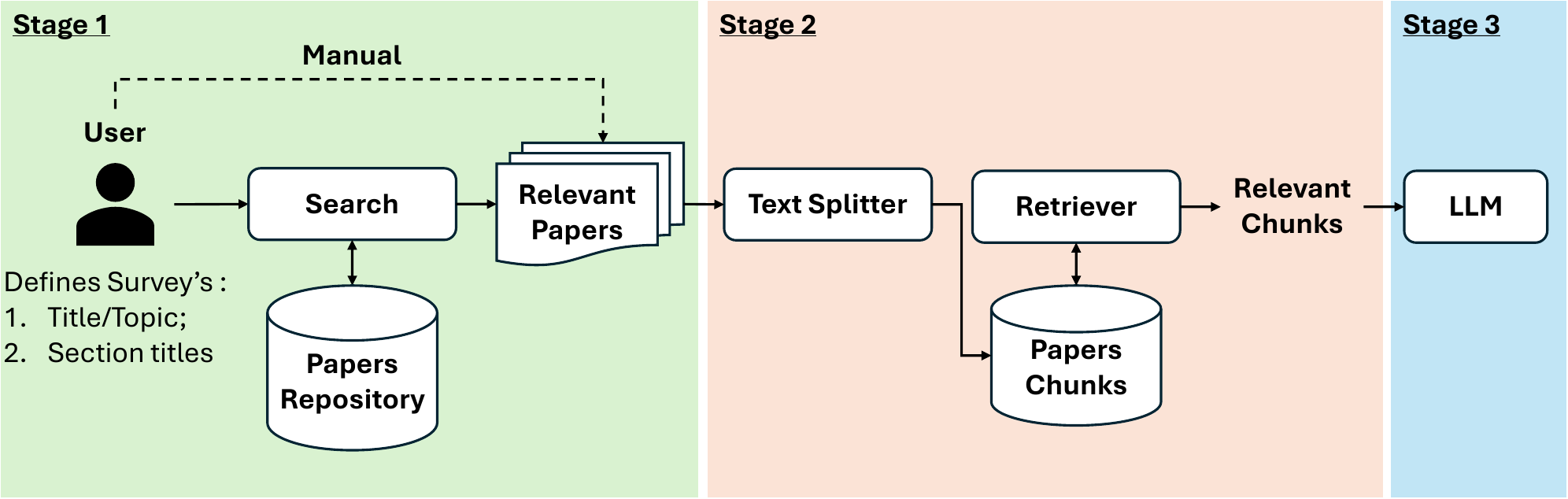}
    \caption{Three stages to generate a section of a survey.}
    \label{fig:general_pipeline}
\end{figure}

The text generation of a survey's section is a multi-document summarization problem. Starting from an input topic, the goal is to find papers related to the topic and summarize them into a single cohesive, coherent, and technically correct text. We present a general pipeline, divided in three stage, to automatically generate a survey's section text from reference papers, as shown in Figure \ref{fig:general_pipeline} and detailed as follows:

\begin{enumerate}
    \item Stage 1: Definition of the title of the survey, the title of the sections, and related papers;
    \item Stage 2: Split the relevant papers texts into chunks and use a search algorithm to retrieve the relevant chunks according to the title of the survey's section;
    \item Stage 3: Generate the text for the section with a Large Language Model (LLM).
\end{enumerate}

The first stage can be done manually, with the researcher choosing a topic/title for the subject and providing a set of pre-selected articles. Or, it can be automated: starting from a topic, one can use a scientific article search engine  (e.g. Semantic Scholar \cite{semantic_scholar}) to find related papers and then refine the results.

The effectiveness of the next stages of the pipeline relies on the quality of the retrieved papers. If the selection is manual, it is assumed that the researcher is aware of which papers are relevant. If the selection is automatic, the quality of the result will be closely tied to the retriever.

The second stage receives the data from the previous stage and selects excerpts from the texts to be consolidated into a single text. This typically includes splitting the text into smaller units, called chunks, storing them in a database, and retrieving the most relevant ones.

Creating the chunks can be done with different strategies. The text can be split  by characters, sentences, or paragraphs. It can involve some degree of overlap or not. The database where the chunks will be stored can be either textual or vector-based. If it's a textual database, the chunks can be enriched using some method (e.g. doc2query \cite{doc2query}). As for a vector database, it's necessary to choose a model to generate the chunk embeddings. Then, for retrieval, one can use a variety of methods -- for example, a lexical search using BM25 \cite{bm25} followed by one or more reranking steps.

Finally, the third stage takes as input the titles of the survey, the section (from stage 1) and the relevant chunks (from stage 2) to summarize using a LLM. To achieve this, it's essential to select the model\footnote{At this stage, there are no restrictions on which LLM to use, but for the purposes of this paper, we experimented only with those from OpenAI.} that will be used.

\subsection{Pipeline 1}

In this pipeline, it is assumed that the subject and title of the section are given to the system, chosen by the researcher. The researcher has the option to provide the papers for each section or let the system search online for relevant papers, in which case the selected articles are downloaded in PDF format and then have their text extracted.

In the second stage, the text from each article are segmented in chunks of 150 words, with an overlap of 50\%. Each chunk is then judged by a neural ranker model, monoT5-3B, which demonstrated a strong zero-shot performance in numerous works~\cite{monot53b}. With the score monoT5-3B gives to each chunk, we select the top $n$ chunks and provide them to the LLM. Figure \ref{fig:prompt_pipeline_1_write_section} shows the prompt template used for generating the content of a section.

\begin{figure}[!htb]
\begin{tcolorbox}[style_prompt]
\textbf{[System]}

You are an expert in the scientific literature review, your job is, given a series of papers and their summaries, write a paragraph with a given title citing relevant information in the traditional format (e.g [1,2,3] [1]) from the provided papers.\\

\textbf{[User]}

Paper 1 \\
Title: Evaluating GPT-3.5 and GPT-4 Models on Brazilian - University Admission Exams \\
Summary: the present study aims to explore the capabilities of Language Models (LMs) in tackling high-stakes multiple-choice tests (...)\\

(...)\\

Paper 5 \\
Title: BLUEX: a benchmark based on Brazilian Leading Universities Entrance eXams \\
Summary: To enable future comparisons, we evaluated our dataset using several language models, ranging from 6B to 66B parameters (...) \\

Paragraph Subject: The capacity and impact of ChatGPT in brazilian education \\

\textbf{[AI]}

Since its introduction, ChatGPT proved that it was capable of handling numerous tasks, [1,2] showed that GPT4 in particular is quite skilled at solving some of the most challenging high school level standardized tests in Brazil, namely ENEM [1], Unicamp and USP [5]...\\

\textbf{[User]}

Paper 1 \\
Title: \{TITLE\_ARTICLE\_1\} \\
Summary: \{SUMMARY\_ARTICLE\_1\} \\
\\
Paper \{N\}\\
Title: \{TITLE\_ARTICLE\_N\}\\
Summary: \{SUMMARY\_ARTICLE\_N\}\\

Paragraph Subject:  \{PARAGRAPH\_SUBJECT\}
\end{tcolorbox}
\caption{Prompt used to write the section of the survey in pipeline 1.}
\label{fig:prompt_pipeline_1_write_section}
\end{figure}

\subsection{Pipeline 2}

In this pipeline, the task is to generate the text for a survey section given the corresponding cited papers associated with it in \proposeddataset.

In the second stage of the general pipeline, each paper's contents are segmented into chunks of 7 sentences with an overlap of 2 sentences to preserve the surrounding context. Then, the embeddings for each chunk are generated using the \specter model \cite{specter2020cohan} and stored in a FAISS vector database \cite{faiss_paper}. \specter was trained to generate embeddings for scientific articles considering the title of the paper and its abstract. In this pipeline, each chunk was encoded considering the title and the chunk.

The retriever selects the 20 most relevant chunks using the section title as the query term. Each chunk is sent to GPT \cite{gpt3}, which provides a score between 0 and 5 using the prompt shown in Figure \ref{fig:prompt_pipeline_2_reranking}.

\begin{figure}[htbp]
\centering
\begin{tcolorbox}[style_prompt]
\textbf{[System]}

You are a renowned scientist who is writing the section `\{TITLE\_SECTION\}' of a survey entitled `\{TITLE\_SURVEY\}'.\\

\textbf{[User]}

I've found a text excerpt from a scientific article that might be useful for the section '\{SECTION\_TITLE\}' of your survey about '\{SURVEY\_TOPIC\}'.\\
Your task is to generate a score for it ranging from 0 to 5 indicating its importance to the section that you are writing.\\
The score of a text written in a language other than English must be 0.\\
If the text contains only metadata of an article (title, authors, venue, etc), i.e. if the text belongs to the references session, the score should be 0.\\
You should also explain why you chose this score.\\
Your answer MUST be enclosed in an RFC8259-compliant JSON object with two properties, "score" and "reasoning", containing the score and the reasoning for it. Remember to enclose the value of the reasoning property in quotes.\\
Do not answer with anything besides the JSON object. Do not insert any text before or after the RFC8259-compliant JSON object.\\
Use the following format to answer (write the reasoning property first and than write the score):\\

\{

\quad"score": \{SCORE\},

\quad"reasoning": \{REASONING\}

\}\\

\textbf{[AI]}

Sure, send me the text I will give you what you need. I will answer with only a JSON object containg the score and the reasoning.\\

\textbf{[User]}

\{CHUNK\_OF\_TEXT\}
\end{tcolorbox}
\caption{Prompt used to rerank the chunks in pipeline 2.}
\label{fig:prompt_pipeline_2_reranking}
\end{figure}

\begin{figure}[htbp]
\begin{tcolorbox}[style_prompt]
\textbf{[System]}

You are a renowned scientist who is writing a survey entitled '\{TITLE\}'.\\

\textbf{[User]}

Your task is to write the text of the section '\{SECTION\_TITLE\}' of the survey. To complete this task, I will give you a list of documents that should be used as references. Each document has a text and an alphanumeric ID.

When writing the section, you MUST follow this rules:

- be aware of plagiarism, i.e., you should not copy the text, but use them as inspiration.

- when using some reference, you must cite it right after its use. You should use the IEEE citing style (write the id of the text between square brackets).

- you are writing the paragraphs of the section. You MUST write only this section.

- you MUST NOT split the section in subsections, nor create introduction and conclusion for it.

- DO NOT write any conclusion in any form for the subsection.

- DO NOT write a references section.

- DO NOT begin the text writing that the context is '{context}', as this is obvious from the title of the survey.

Do you understand your task?\\

\textbf{[AI]}

Sure, send me a list of text and I will write a section about '\{SECTION\_TITLE\}' using them as references. I am aware that I should use the IEEE citing style.\\

\textbf{[User]}

ID: \{REF0\}

Text: \{CHUNK\_0\}\\
...\\
ID: \{REFN\}

Text: \{CHUNK\_N\}\\
\end{tcolorbox}
\caption{Prompt used to write the section of the survey in pipeline 2.}
\label{fig:prompt_pipeline_2_write_section}
\end{figure}

After this step, all chunks that received a score of 0 are discarded. The remaining ones are reordered by the score, and only the top $n$ best-scored are selected and sent to the third stage. These chunks are inserted into the prompt shown in Figure \ref{fig:prompt_pipeline_2_write_section} and sent to GPT to generate the section text.

\FloatBarrier

\subsection{Evaluation Metrics} 

\subsubsection{References F1 Score}
When using LLMs, particularly in the context of Retrieval-Augmented Generation (RAG) systems, it's important to assess the quality of the generated content and the model's effectiveness in using the provided references to generate text. In scientific literature, it is common for each claim to be supported by multiple references. Therefore, one approach is to request the model to produce a summary along with the references it used. This enables the calculation of the F1 score for the references used by the model against the references provided in the input, serving as the ground truth. This method allows for an evaluation of which model or prompt more effectively uses the given information.

\subsubsection{G-eval}

\begin{figure}[!htb]
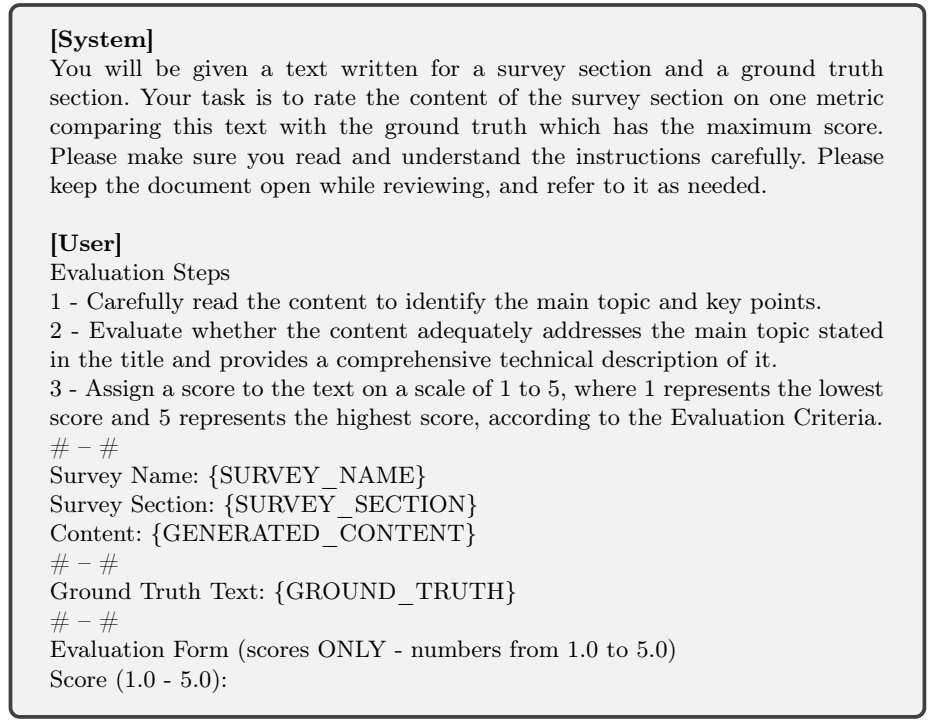

\centering
\begin{tcolorbox}[style_prompt]
\textbf{[System]}

You will be given a text written for a survey section and a ground truth section. Your task is to rate the content of the survey section on one metric comparing this text with the ground truth which has the maximum score. Please make sure you read and understand the instructions carefully. Please keep the document open while reviewing, and refer to it as needed.\\

\textbf{[User]}

Evaluation Steps

1 - Carefully read the content to identify the main topic and key points.

2 - Evaluate whether the content adequately addresses the main topic stated in the title and provides a comprehensive technical description of it.

3 - Assign a score to the text on a scale of 1 to 5, where 1 represents the lowest score and 5 represents the highest score, according to the Evaluation Criteria.

\# -- \#

Survey Name: \{SURVEY\_NAME\}

Survey Section: \{SURVEY\_SECTION\}

Content: \{GENERATED\_CONTENT\}

\# -- \#

Ground Truth Text: \{GROUND\_TRUTH\}

\# -- \#

Evaluation Form (scores ONLY - numbers from 1.0 to 5.0)

Score (1.0 - 5.0):

\end{tcolorbox}
\caption{Prompt used for G-Eval.}
\label{fig:prompt_gpt_score}
\end{figure}

G-eval is a metric for evaluation content generated by LLMs \cite{Liu2023GEvalNE}. It measures the quality by comparing the generated text to a reference text. The final score is calculated by sampling a list of scores provided by an LLM with decoding temperature set to 1, top-$p$ sampling set to 1 and number of samples set to 20. It is defined as follows:
    
    \begin{equation}
        score = \frac{1}{N} \sum_{i=1}^{N} p(s_i) \times s_i
    \end{equation}
    
where \(N\) is the number of samples, \(p(s_i)\) is the probability of the \(i\)-th sample, and \(s_i\) is the score of the \(i\)-th sample. The score is generated using the prompt shown in Figure \ref{fig:prompt_gpt_score}.

\subsubsection{\checkeval}

We also used the \checkeval metric to evaluate the generated text. This metric is based on the methodology proposed by Pereira et al. \cite{checkeval}. The \checkeval metric evaluates the consistency between the candidate text and the reference text using an evaluation checklist. The checklist is generated by a LLM model, which is then used to evaluate the candidate text. The presence of the checklist elements in the candidate text is quantified to determine its information quality. The metric calculates a score reflecting the consistency between the candidate and reference text. For example, if the evaluation checklist consists of ten elements and the candidate text includes eight of these, the \checkeval score will be 0.8, i.e., the metric suggests that the candidate text maintains an 80\% consistency level with the reference text.



\FloatBarrier

\section{Results}

This section presents the results of generating the texts of each section of \proposeddataset using the two pipelines described in Section \ref{sec:pipelines} with different setups.

Pipeline 1 was executed using the monoT5-3B\footnote{\url{https://huggingface.co/castorini/monot5-3b-msmarco}} model to retrieve the chunks and the gpt-3.5-turbo-0125 model to generate the final text. We have evaluated different setups for this pipeline as follows:

\begin{itemize}
    \item 1.1: Section generation with 5 chunks, with the corresponding cited articles from \proposeddataset;
    \item 1.2: Section generation with 10 chunks, with the corresponding cited articles from \proposeddataset;
    \item 1.3: Section generation with 10 chunks, with articles retrieved from the Semantic Scholar API.
\end{itemize}

Pipeline 2 was executed with six different setups after the chunks retriever stage. After selecting the top 20 chunks, a GPT model was used to rerank the chunks and to write the section. Each configuration differs in the GPT model used and the number of chunks considered to summarize the documents. In all of them, only the articles from each section of the dataset were considered as input for the models:

\begin{itemize}
    \item 2.1: gpt-3.5-turbo-0125 model. Section written with a single chunk;
    \item 2.2: gpt-3.5-turbo-0125 model. Section written with up to 5 chunks;
    \item 2.3: gpt-3.5-turbo-0125 model. Section written with up to 10 chunks;
    \item 2.4: gpt-4-0125-preview model. Section written with a single chunk;
    \item 2.5: gpt-4-0125-preview model. Section written with up to 5 chunks;
    \item 2.6: gpt-4-0125-preview model. Section written with up to 10 chunks;
\end{itemize}

Table \ref{tab:results} shows the metrics for nine configurations evaluated. Regarding the metrics used (G-eval and \checkeval), both metrics generally agreed with each other. However, unlike the evaluation checklist, G-eval tends to concentrate the results around certain numbers.

Pipeline 1.3 had the worst results in both \checkeval and G-eval scores among all pipelines. This was the only configuration tested that used articles retrieved from a search engine instead of articles from \proposeddataset. This result shows a strong dependence of text summarization on the retriever stage.

For pipelines of the same type ([1.1 and 1.2], [2.1, 2.2, and 2.3], [2.4, 2.5, and 2.6]), the more chunks used, the higher the F1 score tends to be, as there is a higher probability of selecting chunks from different articles.
However, it cannot be inferred that increasing the number of chunks will necessarily improve text quality. Although the experiments indicated this average behavior for this number of chunks, there are samples where increasing the number of chunks worsened the result. The quality of the chunk — directly influenced by the retriever stage — is very important.

Regardless of the metric considered in Table \ref{tab:results}, the results for pipelines 2.4, 2.5, and 2.6 (using the gpt-4-0125-preview model) were better than the results for pipelines 2.1, 2.2, and 2.3 (equivalent pipelines using gpt-3.5-turbo-0125). This suggests that the LLM used is relevant to the quality of result.

\begin{table}[]
\caption{Results of pipelines. All checklist evaluations were performed using gpt-3-5-turbo and all G-Eval scores were generated by gpt-4.}
\centering
\footnotesize
\resizebox{\textwidth}{!}{%
\begin{tabular}{@{}cccccccc@{}}
\toprule
	
	Pipeline  
	& \begin{tabular}[c]{@{}c@{}}LLM\end{tabular} 
	& \begin{tabular}[c]{@{}c@{}}Number of \\ Chunks\end{tabular}

	& \begin{tabular}[c]{@{}c@{}}Source \end{tabular}

	& \begin{tabular}[c]{@{}c@{}}IR\\ Pipeline\end{tabular}
	& \begin{tabular}[c]{@{}c@{}}\checkeval\\ Score\end{tabular}
	& \begin{tabular}[c]{@{}c@{}}F1\\ Score\end{tabular}
	& \begin{tabular}[c]{@{}c@{}}G-Eval\\ Score\end{tabular} \\ 

\midrule

1.1 & gpt-3.5-turbo-0125 & 5  & \proposeddataset & MonoT5       & \textbf{0.415} & 0.64    & 3.8 \\
1.2 & gpt-3.5-turbo-0125 & 10 & \proposeddataset & MonoT5       & 0.244           & 0.59    & 3.74 \\
1.3 & gpt-3.5-turbo-0125 & 10 & Semantic Scholar & MonoT5       & 0.141           & -       & 2.84 \\
2.1 & gpt-3.5-turbo-0125 & 1  & \proposeddataset & Embeddings   & 0.188           & 0.36   & 3.42 \\
2.2 & gpt-3.5-turbo-0125 & 5  & \proposeddataset & Embeddings   & 0.257           & 0.66    & 3.9 \\
2.3 & gpt-3.5-turbo-0125 & 10 & \proposeddataset & Embeddings   & 0.284           & 0.7     & 3.9 \\
2.4 & gpt-4-0125-preview & 1  & \proposeddataset & Embeddings   & 0.271           & 0.36   & 3.82 \\
2.5 & gpt-4-0125-preview & 5  & \proposeddataset & Embeddings   & 0.323           & 0.67    & 3.99 \\
2.6 & gpt-4-0125-preview & 10 & \proposeddataset & Embeddings   & 0.352           & \textbf{0.72} & \textbf{4.01} \\ \bottomrule
\end{tabular}%
}
\vspace{0.05cm}
\label{tab:results}
\end{table}

The G-Eval and \checkeval scores are high correlated (76\%), indicating consistency in their evaluation of summary quality. However, G-Eval tends to produce results with lower variability compared to the \checkeval score. The \checkeval score, being more granular and item-specific, captures a broader range of evaluation aspects. These insights underscore the importance of employing multiple evaluation metrics to obtain a comprehensive understanding of the performance of summarization models.

\section{Conclusion}

In this paper, we introduced \proposeddataset, a dataset designed for summarizing sections within scientific surveys in artificial intelligence, natural language processing, and machine learning. Our contributions include the creation of the dataset, two pipelines to summarize scientific articles, and their evaluation using various metrics.

As with various other summarization challenges, the summarization of articles for the generation of a survey section does not yield a single definitive answer. Different researchers may use the same set of papers to produce a section on a given topic, each covering different aspects of that topic. Nevertheless, it is important to establish parameters for comparison. \proposeddataset addresses a gap in document summarization by providing a dataset that assists the development and benchmarking of models focused on scientific surveys.

Our results highlight the dependence on the quality of the retriever, as the effectiveness of the summarization pipelines was influenced by the accuracy and relevance of the documents retrieved during the initial stages. High-quality retrieval may ensure that the most relevant information was available for summarization.




\bibliographystyle{plain}
\bibliography{main.bib}

\end{document}